\documentclass[11pt,a4paper]{article}
\usepackage[hyperref]{emnlp-ijcnlp-2019}
\usepackage{times}
\usepackage{latexsym}

\usepackage{url}

\usepackage[pdftex]{graphicx}
\usepackage{paralist}
\usepackage{CJK}
\usepackage{multirow}
\usepackage{subfigure}
\usepackage{algorithm}
\usepackage{algpseudocode}
\usepackage{amsmath}
\usepackage{amssymb}
\usepackage{nccbbb}
\usepackage{array}
\newcolumntype{H}{>{\setbox0=\hbox\bgroup}c<{\egroup}@{}}
\newcolumntype{Z}{>{\setbox0=\hbox\bgroup}c<{\egroup}@{\hspace*{-\tabcolsep}}}

\aclfinalcopy 

\title{DyKgChat: Benchmarking Dialogue Generation\\ Grounding on Dynamic Knowledge Graphs}

\author{Yi-Lin Tuan \quad Yun-Nung Chen \quad Hung-yi Lee\\
  National Taiwan University, Taipei, Taiwan \\
  {\tt pascaltuan@gmail.com\quad y.v.chen@ieee.org\quad hungyilee@ntu.edu.tw}
  \\}

\date{}

\begin{document}
\maketitle
\begin{abstract}
  Data-driven, knowledge-grounded neural conversation models are capable of generating more informative responses.
  However, these models have not yet demonstrated that they can zero-shot adapt to updated, unseen knowledge graphs.
  This paper proposes a new task about how to apply dynamic knowledge graphs in neural conversation model and presents a novel TV series conversation corpus (DyKgChat) for the task.
  Our new task and corpus aids in understanding the influence of dynamic knowledge graphs on responses generation.
  Also, we propose a preliminary model that selects an output from two networks at each time step: a sequence-to-sequence model (Seq2Seq) and a multi-hop reasoning model, in order to support dynamic knowledge graphs.
  To benchmark this new task and evaluate the capability of adaptation, we introduce several evaluation metrics and the experiments show that our proposed approach outperforms previous knowledge-grounded conversation models.
  The proposed corpus and model can motivate the future research directions\footnote{The data and code are available in \url{https://github.com/Pascalson/DyKGChat}.}.
\end{abstract}

\section{Introduction}

\begin{figure}[t!]
    \centering
    \includegraphics[width=\linewidth]{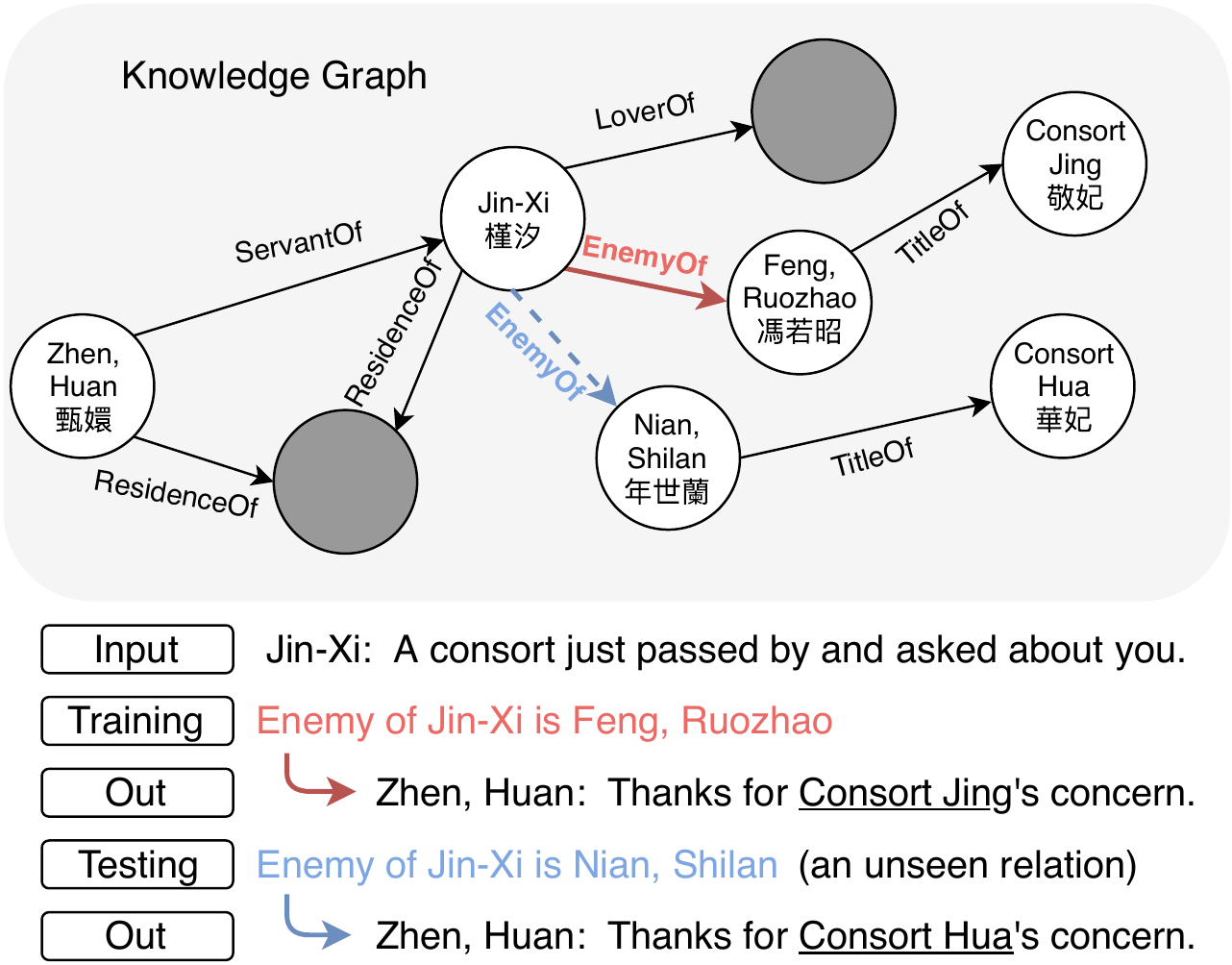}
    \caption{An example of an ideal conversation model with dynamic knowledge graphs.}
    \label{fig:intro}
    \vspace{-3mm}
\end{figure}

In the chit-chat dialogue generation, neural conversation models~\cite{sutskever2014sequence, sordoni2015neural, vinyals2015neural} have emerged for its capability to be fully data-driven and end-to-end trained.
While the generated responses are often reasonable but \emph{general} (without useful information), recent work proposed knowledge-grounded models~\cite{eric2017key, ghazvininejad2018knowledge, zhou2018commonsense, qian2018assigning} to incorporate external facts in an end-to-end fashion without hand-crafted slot filling.
Effectively combining text and external knowledge graphs have also been a crucial topic in \emph{question answering}~\cite{yin2016neural, hao2017end, levy2017zero, sun2018open, das2018building}.
Nonetheless, prior work rarely analyzed the model capability of zero-shot adaptation to dynamic knowledge graphs,
where the states/entities and their relations are temporal and evolve as a single time scale process.
For example, as shown in Figure~\ref{fig:intro}, the entity \emph{Jin-Xi} was originally related to the entity \emph{Feng, Ruozhao} with the type \emph{EnemyOf}, but then evolved to be related to the entity \emph{Nian, Shilan}.

\begin{CJK*}{UTF8}{bkai}
\begin{table*}[t!]\small
    \centering
    \begin{tabular}{cl}\hline
        \multirow{4}{*}{HGZHZ} & {\bf Zhen-Huan}: She must be frightened. It should blame me. I should not ask her to play chess.\\
        & {\bf 甄嬛}: 姐姐定是嚇壞了。都怪臣妾不好，好端端的來叫梅姐姐下棋做什麼。\\
        & {\bf Doctor-Wen}: Relax, {\bf Concubine-Huan} {\bf madame}. {\bf Lady Shen} is just injured but is fine.\\
        & {\bf 溫太醫}: {\bf 莞嬪}{\bf 娘娘}請放心，{\bf 惠貴人}的精神倒是沒有大礙，只是傷口燒得有些厲害。\\\hline
        \multirow{2}{*}{Friends} & {\bf Joey}: C' mon , you're going out with the guy! There's gotta be something wrong with him!\\
        & {\bf Chandler}: Alright {\bf Joey}, be nice. So does he have a hump? A hump and a hairpiece?\\\hline
    \end{tabular}
    \caption{Examples of DyKgChat corpus.}
    \label{tab:data-example}
    \vspace{-2mm}
\end{table*}
\end{CJK*}

The goal of this paper is to facilitate knowledge-grounded neural conversation models to learn and zero-shot adapt with dynamic knowledge graphs.
To our observation, however, there is no existing conversational data paired with dynamic knowledge graphs.
Therefore, we collect a TV series corpus---\emph{DyKgChat}, with facts of the fictitious life of characters.
DyKgChat includes a Chinese palace drama \emph{Hou Gong Zhen Huan Zhuan (HGZHZ)}, and an English sitcom \emph{Friends},
which contain dialogues, speakers, scenes (e.g., the places and listeners), and the corresponded knowledge graphs including explicit information such as the relations \emph{FriendOf}, \emph{EnemyOf}, and \emph{ResidenceOf} as well as the linked entities.
Table~\ref{tab:data-example} shows some examples from DyKgChat.

Prior graph embedding based knowledge-grounded conversation models~\cite{sutskever2014sequence,ghazvininejad2018knowledge,zhu2017flexible} did not directly use the graph structure, so it is unknown how a changed graph will influence the generated responses.
In addition, key-value retrieved-based models~\cite{yin2016neural, eric2017key, levy2017zero, qian2018assigning} retrieve only \emph{one-hop} relation paths.
As fictitious life in drama, realistic responses often use knowledge entities existing in multi-hop relational paths, e.g., the residence of a friend of mine.
Therefore, we propose a model that incorporates multi-hop reasoning~\cite{lao2011random, neelakantan2015compositional, xiong2017deeppath} on the graph structure into a neural conversation generation model.
Our proposed model, called quick adaptive dynamic knowledge-grounded neural conversation model (Qadpt), is based on a Seq2Seq model~\cite{sutskever2014sequence} with a widely-used copy mechanism~\cite{gu2016incorporating, merity2016pointer, he2017generating, xing2017topic, zhu2017flexible, eric2017key, ke2018generating}.
To enable multi-hop reasoning, the model factorizes a transition matrix for a \emph{Markov chain}.

In order to focus on the capability of producing reasonable knowledge entities and adapting with dynamic knowledge graphs, we propose two types of automatic metrics.
First, given the provided knowledge graphs, we examine if the models can generate responses with proper usage of multi-hop reasoning over knowledge graphs.
Second, after randomly replacing some crucial entities in knowledge graphs, we test if the models can accordingly generate correspondent responses.
The empirical results show that our proposed model has the desired advantage of zero-shot adaptation with dynamic knowledge graphs, and can serve as a preliminary baseline for this new task.
To sum up, the contributions of this paper are three-fold:
\begin{compactitem}
\item A new task, \emph{dynamic knowledge-grounded conversation generation}, is proposed.
\item A newly-collected TV series conversation corpus \emph{DyKgChat} is presented for the target task.
\item We benchmark the task by comparing many prior models and the proposed quick adaptive dynamic knowledge-grounded neural conversation model (Qadpt), providing the potential of benefiting the future research direction.
\end{compactitem}

\section{Task Description}
For each single-turn conversation, the input message and response are respectively denoted as $x=\{x_t\}_{t=1}^{m}$ and $y=\{y_t\}_{t=1}^{n}$, where $m$ and $n$ are their lengths.
Each turn $(x, y)$ is paired with a knowledge graph $\mathcal{K}$, which is composed of a collection of triplets $(h, r, t)$, where $h,t\in \mathcal{V}$ (the set of entities) and $r\in \mathcal{L}$ (the set of relationships).
Each word $y_t$ in a response belongs to either generic words $\mathcal{W}$ or knowledge graph entities $\mathcal{V}$.
The task is two-fold:
\begin{compactenum}
\item Given an input message $x$ and a knowledge graph $\mathcal{K}$, the goal is to generate a sequence $\{\hat{y}_t\}_{t=1}^{n}$ that is not only as similar as possible to the ground-truth $\{y_t\}_{t=1}^{n}$, but contains \emph{correct knowledge graph entities} to reflect the information.
\item After a knowledge graph is updated to $\mathcal{K}'$, where some triplets are revised to $(h,r,t')$ or $(h,r',t)$, the generated sequence should contain \emph{correspondent knowledge graph entities} in $\mathcal{K}'$ to reflect the updated information.
\end{compactenum}

\subsection{Evaluation Metrics}
To evaluate dynamic knowledge-grounded conversation models, we propose two types of evaluation metrics for validating two aspects described above.

\subsubsection{Knowledge Entity Modeling}
There are three metrics focusing on the knowledge-related capability.
\label{sec:eval-1}
\paragraph{Knowledge word accuracy (KW-Acc).}
Given the ground-truth sentence as the decoder inputs, at each time step, it evaluates how many knowledge graph entities are correctly predicted.
\begin{align}
    \text{KW-Acc} = \sum_{t=1}^{n} & P(\hat{y}_t = y_t \mid y_1 y_2 \dots y_{t-1}, y_t\in \mathcal{V}).\nonumber
\end{align}
For example, after perceiving the partial ground-truth response ``\emph{If Jin-Xi not in}'' and knowing the next word should be a knowledge graph entity, KW-Acc measures if the model can predict the correct word ``\emph{Yongshou Palace}''.
\paragraph{Knowledge and generic word classification (KW/Generic).}
Given the ground-truth sentence as the decoder inputs, at each time step, it measures the capability of predicting the correct class (a knowledge graph entity or a generic word) and adopts micro-averaging.
The true positive, false negative and false positive are formulated as:
\begin{align}
    \mathrm{TP} &= |\{t\mid \hat{y}_t\in \mathcal{V}, y_t\in \mathcal{V}\}|,\nonumber\\
    \mathrm{FN} &= |\{t\mid \hat{y}_t\in \mathcal{W}, y_t\in \mathcal{V}\}|,\nonumber\\
    \mathrm{FP} &= |\{t\mid \hat{y}_t\in \mathcal{V}, y_t\in \mathcal{W}\}|,\nonumber\\
    \hat{y}_t &\sim P(\cdot\mid y_1 y_2 \dots y_{t-1}).\nonumber
\end{align}
\paragraph{Generated knowledge words (Generated-KW).}
Considering the knowledge graph entities in the reference $y=\{y_t\}_{t=1}^{n}$ as positives,
in the inference stage, we use the generated knowledge entities to compute true positive, false positive, and true negative, and adopt micro-averaging.
\begin{align}
    \mathrm{TP} &= |\{ \hat{y}_t\in \{y_t\in \mathcal{V}\}_{t=1}^{n}, \hat{y}_t\in \mathcal{V}\}_{t=1}^{n}|,\nonumber\\
    \mathrm{FN} &= |\{ y_t\notin \{\hat{y}_t\in \mathcal{V}\}_{t=1}^{n}, y_t\in \mathcal{V}\}_{t=1}^{n}|,\nonumber\\
    \mathrm{FP} &= |\{ \hat{y}_t\notin \{y_t\in \mathcal{V}\}_{t=1}^{n}, \hat{y}_t\in \mathcal{V}\}_{t=1}^{n}|,\nonumber\\
    \hat{y}_t &\sim P(\cdot\mid \hat{y}_1 \hat{y}_2 \dots \hat{y}_{t-1}).\nonumber
\end{align}
For example, after input a sentence ``\emph{Where's {\bf JinXi}?}'', if a model generates ``\emph{Hi, {\bf Zhen-Huan}, {\bf JinXi} is in {\bf Yangxin-Palace}.}'' when reference is ``\emph{{\bf JinXi} is in {\bf Yongshou-Palace}.}'', where bolded words are knowledge entities. Recall is $\frac{1}{2}$ and precision is $\frac{1}{3}$.

\subsubsection{Adaptation of Changed Knowledge Graphs}
\label{sec:eval-2}
Each knowledge graph is randomly changed by (1) shuffling a batch (\emph{All}), (2) replacing the predicted entities (\emph{Last1}), or (3) replacing the last two steps of paths predicting the generated entities (\emph{Last2}).
We have two metrics focusing on the capability of adaptation.
\paragraph{Change rate.}
It measures if the responses are different from the original ones (with original knowledge graphs).
The higher rate indicates that the model is more sensitive to a changed knowledge graph.
Therefore, the higher rate may not be better, because some changes are worse. The following metric is proposed to deal with the issue, but \emph{change rate} is still reported.
\paragraph{Accurate change rate.}
This measures if the original predicted entities are replaced with the hypothesis set, where this ensures that the updated responses generate knowledge graph entities according to the updated knowledge graphs.
(1) In \emph{All}, the hypothesis set is the collection of all entities in the new knowledge graph.
(2) In \emph{Last1} and \emph{Last2}, the hypothesis set is the randomly-selected substitutes.

\begin{table}[t!]\small
    \centering
    \begin{tabular}{lrr}\hline
        \bf Metrics &  \bf HGZHZ & \bf Friends\\\hline
        \# Dialogues & 1247 & 3092 \\
        Total \# turns & 17,164 & 57,757 \\
        Total \# tokens & 462,647 & 838,913 \\
        Avg. turns per dialogue & 13.76 & 18.68 \\
        Avg. tokens per turn & 26.95 & 14.52 \\
        Total unique tokens & 3,624 & 19,762 \\
        \# KG entities & 174 & 281 \\
        \# KG relation types & 9 & 7 \\
        total \# KG entities appear & 46,059 & 176,550 \\
        \# Dialogues w/ KG entities & 1,166 & 2,373 \\
        \# turns w/ KG entities & 10,110 & 9,199 \\\hline
    \end{tabular}
    \caption{The details of collected DyKgChat.}
    \label{tab:data-details}
\end{table}
\begin{table}[t!]\small
    \centering
    \begin{tabular}{cl}\hline
         & \bf Relation Type (Percentage)\\\hline
        \multirow{5}{*}{\bf HGZHZ} & IsAliasOf (25\%), IsChildOf (5\%), \\ & IsLoverOf (6\%), IsParentOf (5\%), \\
        & IsResidenceOf (16\%), IsSiblingOf (2\%),\\
        & IsTitleOf (30\%), IsEnemyOf (8\%),\\
        & IsServantOrMasterOf (3\%)\\\hline
        \multirow{4}{*}{\bf Friends} & IsLoverOf (12\%), IsWorkplaceOf (2\%),\\
        & IsOccupationOf (8\%), IsNameOf (47\%),\\
        & IsRelativeOf (8\%), IsFriendOf (4\%),\\
        & IsNicknameOf (19\%)\\\hline
    \end{tabular}
    \caption{The included relation types in the collect knowledge graphs, and their percentages.}
    \label{tab:relation-types}
\end{table}

\begin{figure*}[t!]
    \centering
    \subfigure[The word cloud of HGZHZ.]{
        \includegraphics[width=.48\linewidth, trim={0.5cm 2.0cm 1cm 0.5cm},clip]{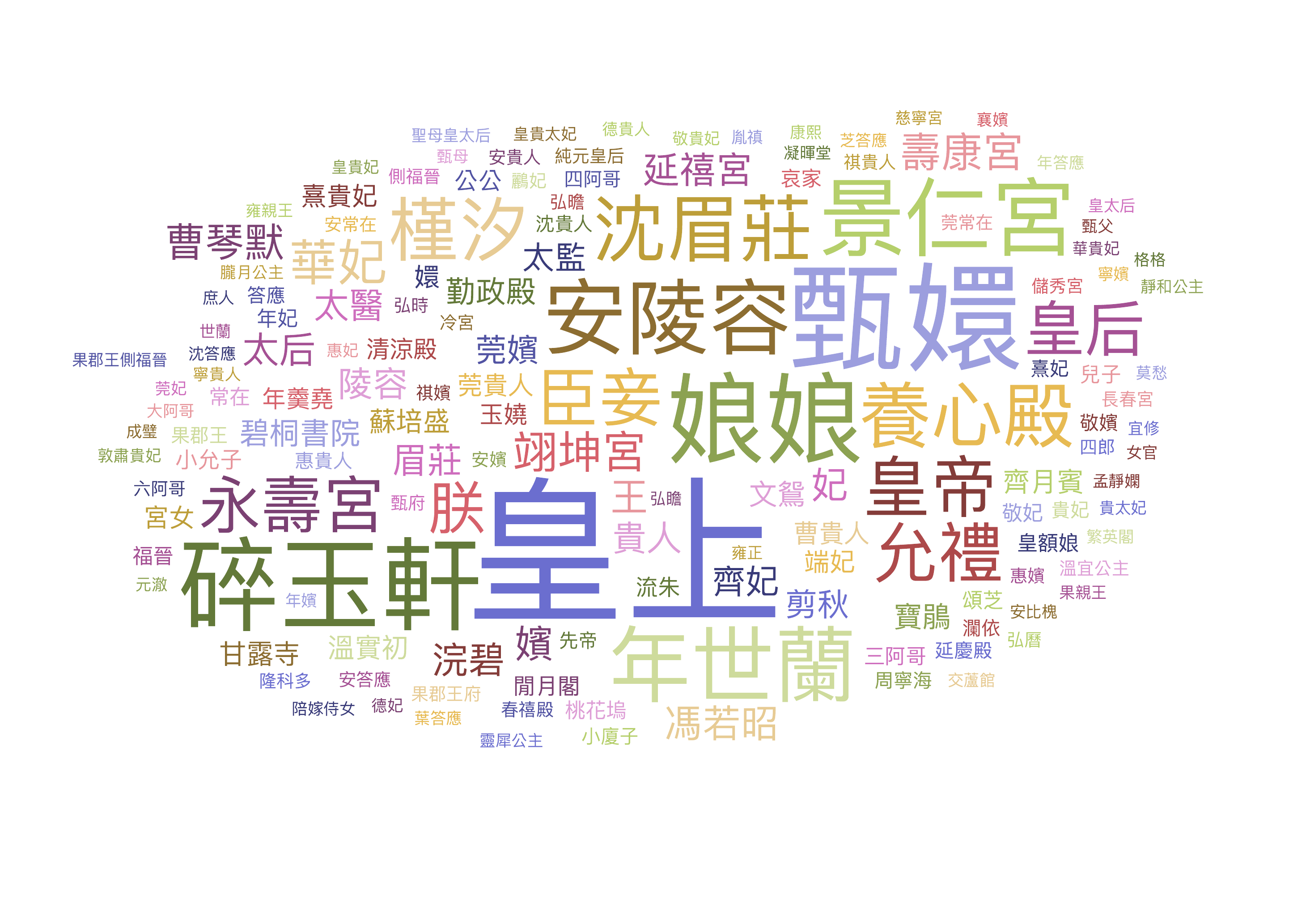}
    }
    \subfigure[The word cloud of Friends.]{
        \includegraphics[width=.48\linewidth, trim={0.5cm 1.5cm 1cm 1.5cm},clip]{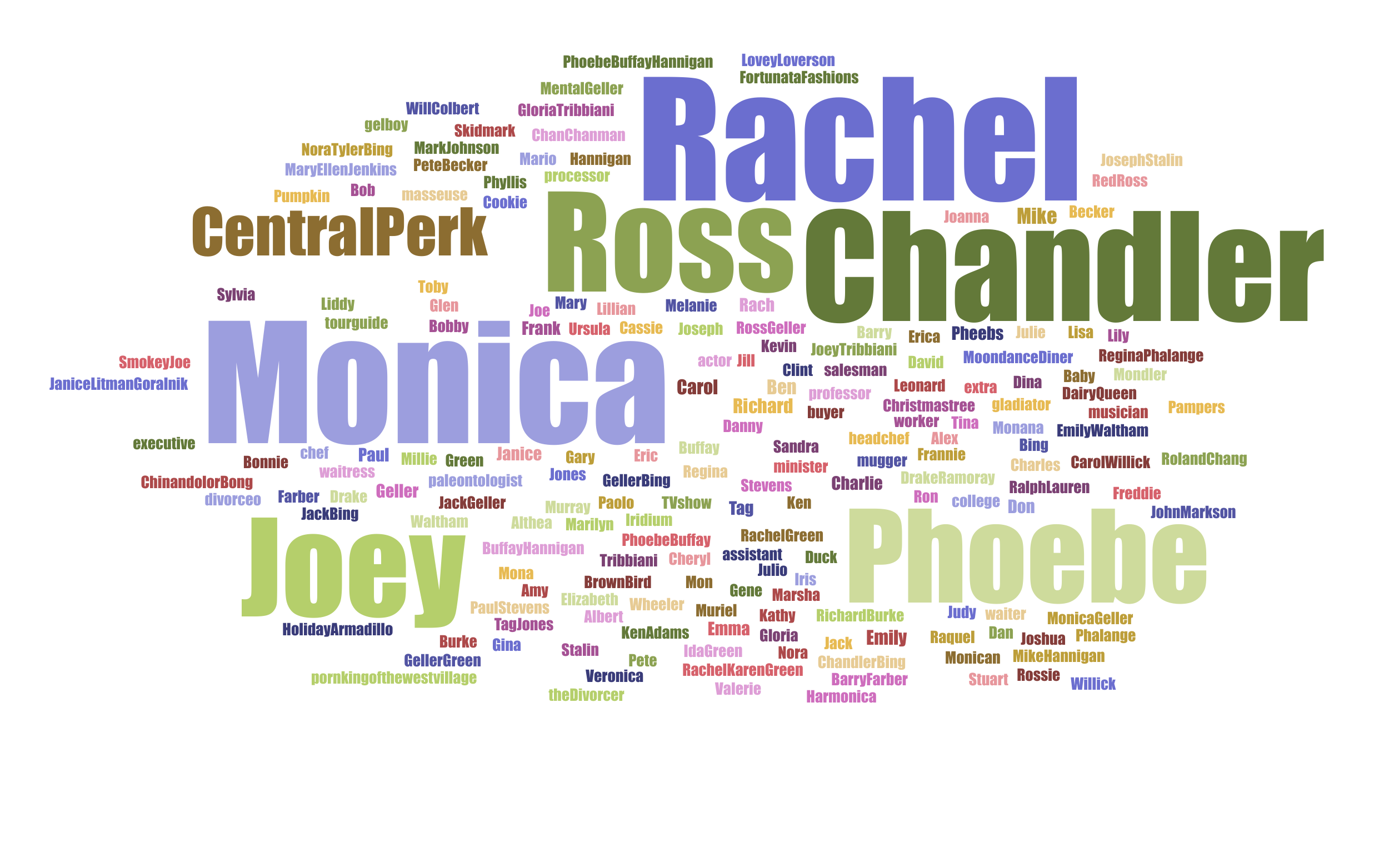}
    }
    \vspace{-2mm}
    \caption{The knowledge entity counts of HGZHZ are more balanced than ones of Friends.}
    \label{fig:wordcloud}
    \vspace{-2mm}
\end{figure*}

\section{DyKgChat Corpus}
This section introduces the collected DyKgChat corpus for the target knowledge-grounded conversation generation task.

\subsection{Data Collection}
To build a corpus where the knowledge graphs would naturally evolves, we collect TV series conversations, considering that TV series often contain complex relationship evolution, such as friends, jobs, and residences.
We choose TV series with different languages and longer episodes.
We download the scripts of a Chinese palace drama ``\emph{Hou Gong Zhen Huan Zhuang}'' (HGZHZ; with 76 episodes and hundreds of characters) from \emph{Baidu Tieba},
and the scripts of an English sitcom ``\emph{Friends}'' (with 236 episodes and six main characters)\footnote{\url{https://github.com/npow/friends-chatbot}}.
Their paired knowledge graphs are manually constructed according to their wikis written by fans\begin{CJK*}{UTF8}{bkai}\footnote{\url{https://zh.wikipedia.org/wiki/}後宮甄嬛傳\_(電視劇)}\end{CJK*}\footnote{\url{https://friends.fandom.com/wiki/Friends_Wiki}}.
Noted that the knowledge graph of HGZHZ is mainly built upon the top twenty-five appeared characters.

The datasets are split 5\% as validation data and 10\% as testing data, where the split is based on multi-turn dialogues and balanced on speakers.
The boundaries of dialogues are annotated in the original scripts.
The tokenization of HGZHZ considers Chinese characters and knowledge entities;
the tokenization of Friends considers space-separated tokens and knowledge entities.
The data statistics after preprocessing is detailed in Table~\ref{tab:data-details}.
The relation types $r\in \mathcal{L}$ of each knowledge graph and their percentages are listed in Table~\ref{tab:relation-types}, and the knowledge graph entities are plotted as word clouds in Figure~\ref{fig:wordcloud}.

\subsection{Subgraph Sampling}
\label{sec:subgraph-sampling}
Due to the excessive labor of building dynamic knowledge graphs aligned with all episodes, we currently collect a fixed knowledge graph $\mathcal{G}$ containing all information that once exists for each TV series.
To build the aligned dynamic knowledge graphs, we sample the top-five shortest paths on knowledge graphs from each source to each target, where the sources are knowledge entities in the input message and the scene $\{x_t\in \mathcal{V}\}$, and the targets are knowledge entities in the ground-truth response $\{y_t\in \mathcal{V}\}$.
We manually check whether the selected number of shortest paths are able to cover most of the used relational paths.
The dynamic knowledge graphs are built based on an ensemble of the following possible subgraphs:
\begin{compactitem}
    \item The sample for each single-turn dialogue.
    \item The sample for each multi-turn dialogue.
    \item The manually-annotated subgraph for each period.
\end{compactitem}
While the first rule is adopted for simplicity, the preliminary models should at least work on this type of subgraphs.
The subgraphs are defined as the dynamic knowledge graphs $\{\mathcal{K}\}$, which are updated every single-turn dialogue.

\begin{figure}[t!]
    \centering
    \includegraphics[width=.9\linewidth]{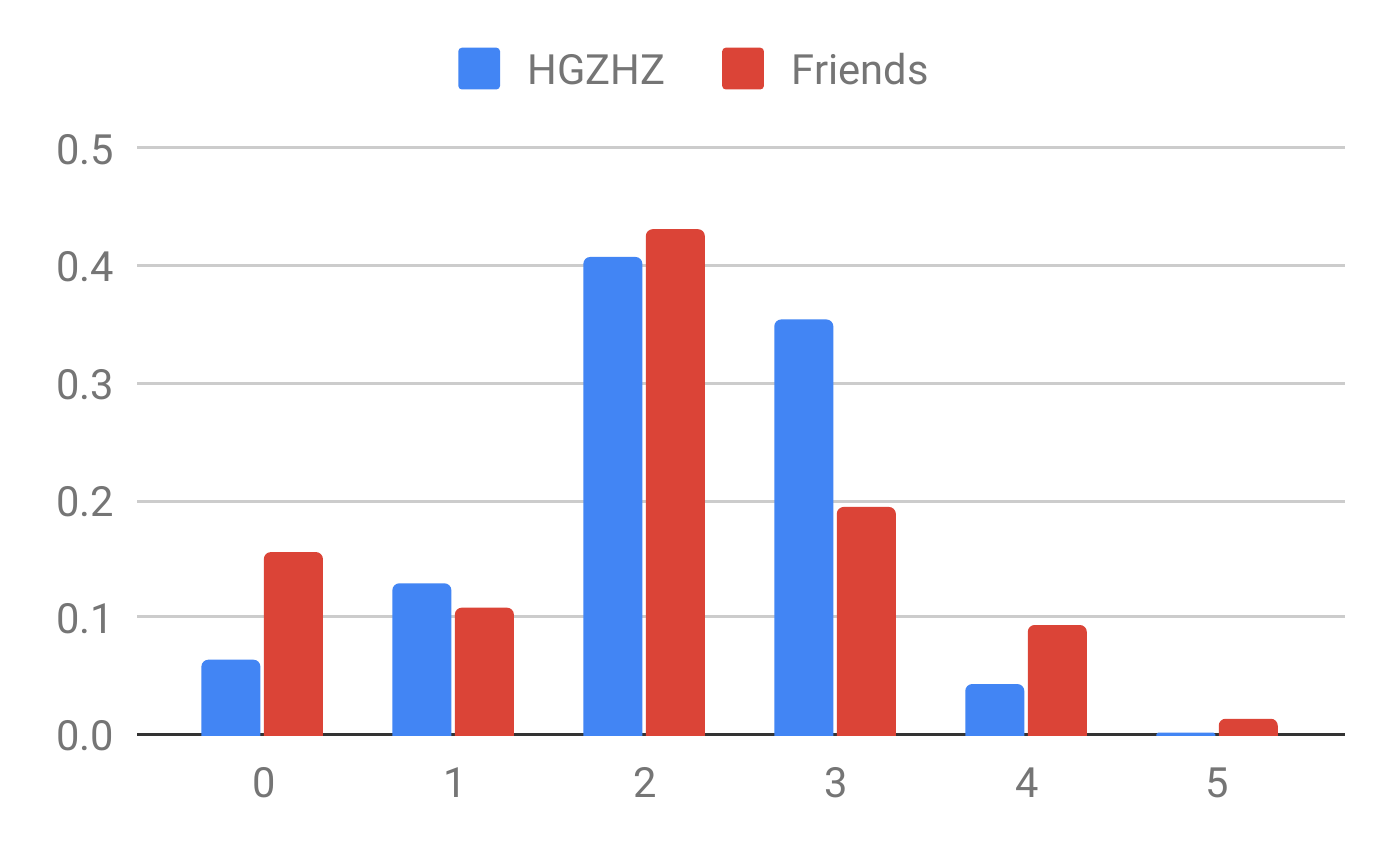}
    \vspace{-2mm}
    \caption{The distribution of lengths of shortest paths.}
    \label{fig:sssp}
    \vspace{-2mm}
\end{figure}

\begin{figure*}[t!]
    \centering
    \includegraphics[width=.95\linewidth]{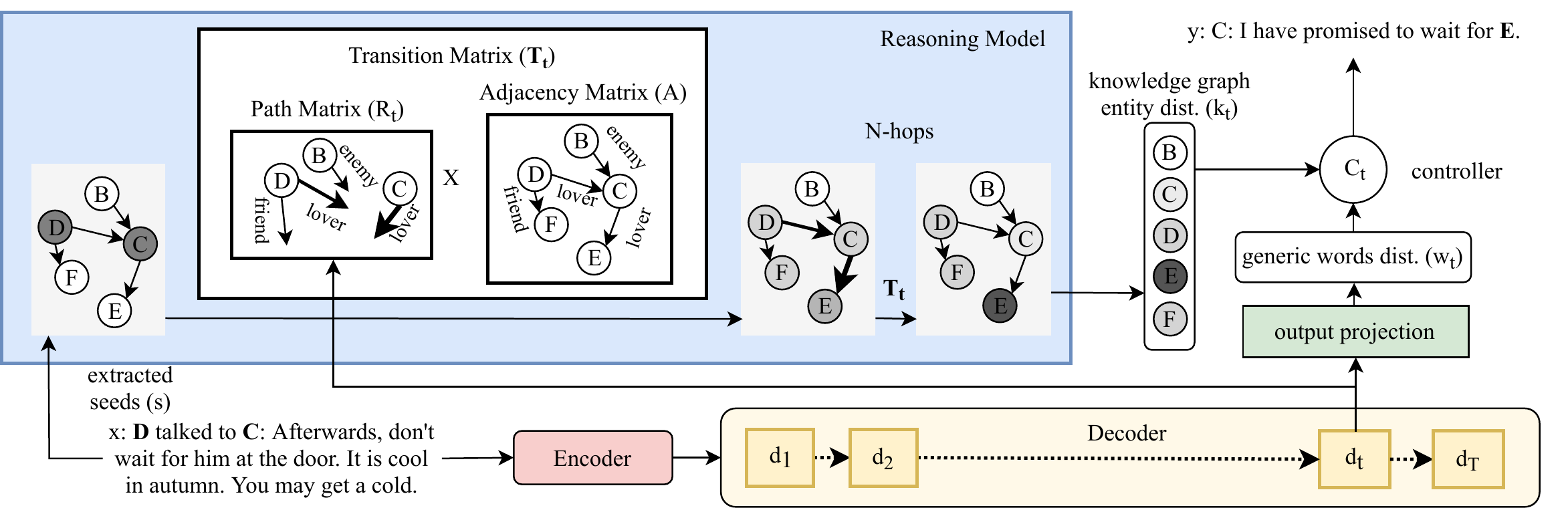}
    \caption{The framework of the proposed model. The node E here is the symbol for the emperor.}
    \label{fig:model}
    \vspace{-2mm}
\end{figure*}

\subsection{Data Analysis}
\paragraph{Data imbalance.}
As shown in Table~\ref{tab:data-details}, the turns with knowledge graph entities are about $58.9\%$ and $15.93\%$ of HGZHZ and Friends respectively.
Apparently in Friends, the training data with knowledge graph entities are too small, so fine-tuning on this subset with knowledge graph entities might be required.


\paragraph{Shortest paths.} The lengths of shortest paths from sources to targets are shown in Figure~\ref{fig:sssp}. 
Most probabilities lie on two and three hops rather than zero and one hop, so key-value extraction based text generative models~\cite{eric2017key, levy2017zero, qian2018assigning} are not suitable for this task.
On the other hand, multi-hop reasoning might be useful for better retrieving correct knowledge graph entities.

\paragraph{Dynamics.} The distribution of graph edit distances among dynamic knowledge graphs are $57.24\pm24.34$ and $38.16\pm15.99$ for HGZHZ and Friends respectively, revealing that the graph dynamics are spread out: some are slightly changed while some are largely changed, which matches our intuition.

\section{Qadpt: Quick Adaptative Dynamic Knowledge-Grounded Neural Conversation Model}
\label{sec:Qadpt}

To our best knowledge, no prior work focused on dynamic knowledge-grounded conversation; thus we propose \emph{Qadpt} as the preliminary model.
As illustrated in Figure~\ref{fig:model}, the model is composed of (1) a Seq2Seq model with a controller, which decides to predict knowledge graph entities $k\in \mathcal{V}$ or generic words $w\in \mathcal{W}$, and (2) a reasoning model, which retrieves the relational paths in the knowledge graph.


\subsection{Sequence-to-Sequence Model}
Qadpt is based on a Seq2Seq model~\cite{sutskever2014sequence, vinyals2015neural}, where the encoder encodes an input message $x$ into a vector $\mathbf{e}(x)$ as the initial state of the decoder.
At each time step $t$, the decoder produces a vector $\mathbf{d}_t$ conditioned on the ground-truth or predicted $y_1 y_2\dots y_{t-1}$.
Note that we use gated recurrent unit (GRU)~\cite{cho2014properties} in our experiments.
\begin{align}
    \mathbf{e}(x) & = \mathbf{GRU}(x_1 x_2 \dots x_m)\\
    \mathbf{d_t} & = \mathbf{GRU}(y_1 y_2\dots y_{t-1}, \mathbf{e}(x))
\end{align}
Each predicted $\mathbf{d}_t$ is used for three parts: 
\emph{output projection}, \emph{controller}, and \emph{reasoning}.
For output projection, the predicted $\mathbf{d}_t$ is transformed into a distribution $\mathbf{w}_t$ over generic words $\mathcal{W}$ by a projection layer.

\subsection{Controller}

To decide which vocabulary set (knowledge graph entities $\mathcal{V}$ or generic words $\mathcal{W}$) to use, the vector $\mathbf{d}_t$ is transformed to a controller $c_t$, which is a widely-used component~\cite{eric2017key, zhu2017flexible, ke2018generating, zhou2018commonsense, xing2017topic} similar to copy mechanism~\cite{gu2016incorporating, merity2016pointer, he2017generating}.
The controller $c_t$ is the probability of choosing from knowledge graph entities $\mathcal{V}$, while $1-c_t$ is the probability of choosing from generic words $\mathcal{W}$.
Note that we take the controller as a special symbol $\mathit{KB}$ in generic words, so the term $1-c_t$ is already multiplied to $\mathbf{w}_t$.
The controller here can be flexibly replaced with any other model.
\begin{equation}
\begin{split}
    P(\{\mathit{KB}, \mathcal{W}\}\mid y_1 y_2\dots &y_{t-1}, \mathbf{e}(x))\\
        & = \textrm{softmax}(\phi(\mathbf{d}_t)),\\
\end{split}
\end{equation}\vspace{-22pt}
\begin{align}
    \mathbf{w}_t & = P(\mathcal{W}\mid y_1 y_2\dots y_{t-1}, \mathbf{e}(x)),\\
    c_t & = P(\mathit{KB} \mid y_1 y_2\dots y_{t-1}, \mathbf{e}(x)),\\
    \mathbf{o}_t & = \{c_t\mathbf{k}_t; \mathbf{w}_t\},
\end{align}
where $\phi$ is the output projection layer, and $\mathbf{k}_t$ is the predicted distribution over knowledge graph entities $\mathcal{V}$ (detailed in subsection~\ref{sec:reasoning-model}), and $\mathbf{o}_t$ is the produced distribution over all vocabularies.

\subsection{Reasoning Model}
\label{sec:reasoning-model}
To ensure that Qadpt can zero-shot adapt to dynamic knowledge graphs, instead of using attention mechanism on graph embeddings~\cite{ghazvininejad2018knowledge, zhou2018commonsense}, we leverage the concept of multi-hop reasoning~\cite{lao2011random}.
The reasoning procedure can be divided into two stages: (1) forming a transition matrix and (2) reasoning multiple hops by a Markov chain.

In the first stage, a transition matrix $\mathbf{T}_t$ is viewed as  multiplication of a path matrix $\mathbf{R}_t$ and the adjacency matrix $\mathbf{A}$ of a knowledge graph $\mathcal{K}$.
The adjacency matrix is a binary matrix indicating if the relations between two entities exist. 
The path matrix is a linear transformation $\theta$ of $\mathbf{d}_t$, and represents the probability distribution of each head $h\in\mathcal{V}$ choosing each relation type $r\in\mathcal{L}$.
Note that a relation type \emph{self-loop} is added.
\begin{align}
    \mathbf{R}_t & = \textrm{softmax}(\theta(\mathbf{d}_t)),\\
    \mathbf{A}_{i, j, \gamma} & = \left\{
        \begin{array}{l}
            1, \quad (h_i, r_j, t_\gamma)\in \mathcal{K}\\
            0, \quad (h_i, r_j, t_\gamma)\notin \mathcal{K}
        \end{array}
        \right.,\\
    \mathbf{T}_t & = \mathbf{R}_t \mathbf{A},
\end{align}
where $\mathbf{R}_t \in \bbbr^{\left |\mathcal{V}\right |\times 1\times\left |\mathcal{L}\right |}$,
$\mathbf{A} \in \bbbr^{\left |\mathcal{V}\right |\times \left |\mathcal{L}\right |\times \left |\mathcal{V}\right |}$,
and $\mathbf{T}_t \in \bbbr^{\left |\mathcal{V}\right |\times\left |\mathcal{V}\right |}$.

In the second stage, a binary vector $\mathbf{s}\in \bbbr^{\left |\mathcal{V}\right |}$ is computed to indicate whether each knowledge entity exists in the input message $x$.
First, the vector $\mathbf{s}$ is multiplied by the transition matrix.
A new vector $\mathbf{s}^\intercal\mathbf{T}_t$ is then produced to denote the new probability distribution over knowledge entities after one-hop reasoning.
After $N$ times reasoning\footnote{We choose $N=6$ because of the maximum length of shortest paths in Figure~\ref{fig:sssp}}, the final probability distribution over knowledge entities is taken as the generated knowledge entity distribution $\mathbf{k}_t$:
\begin{align}
    \mathbf{k}_t & = \mathbf{s}^\intercal (\mathbf{T}_t)^N.
\end{align}
The loss function is the cross-entropy of the predicted word $\mathbf{o}_t$ and the ground-truth distribution:
\begin{gather}
    \mathcal{L}(\psi, \phi, \theta) = -\sum_{t=1}^{n} \log \mathbf{o}_t(y_t),
\end{gather}
where $\psi$ is the parameters of GRU layers.
Compared to prior work, the proposed reasoning approach explicitly models the knowledge reasoning path, so an updated knowledge graphs will definitely change the results without retraining.

\subsection{Inferring Reasoning Paths}
Because this reasoning method is stochastic,
we compute the probabilities of the possible reasoning paths by the reasoning model, and infer the one with the largest probability as the retrieved reasoning path.

\section{Related Work}
The proposed task is motivated by prior knowledge-grounded conversation tasks \cite{ghazvininejad2018knowledge, zhou2018commonsense}, but further requires the capability to adapt to dynamic knowledge graphs.

\begin{table*}[t!]\small
    \centering
    \begin{tabular}{l|c|c|c|c|c|c|c|c|c|c|c|c}\hline
        \multirow{3}{*}{\bf Model} 
        & \multicolumn{6}{c|}{\bf HGZHZ}
        & \multicolumn{6}{c}{\bf Friends}\\\cline{2-13}
        & \multicolumn{3}{c|}{\bf Change Rate}
        & \multicolumn{3}{c}{\bf Accurate Change Rate}
        & \multicolumn{3}{|c|}{\bf Change Rate}
        & \multicolumn{3}{c}{\bf Accurate Change Rate}\\\cline{2-13}
        & \bf All & \bf Last1 & \bf Last2
        & \bf All & \bf Last1 & \bf Last2
        & \bf All & \bf Last1 & \bf Last2
        & \bf All & \bf Last1 & \bf Last2\\\hline
        MemNet & 92.98 & 31.78 & 37.46
        & 62.19 & 1.17 & 2.92
        & 93.27 & 19.22 & 27.53
        & 78.23 & 0.91 & 7.66\\
        ~~+ multi & 98.69 & 77.87 & 81.96
        & \bf 83.82 & 3.40 & 10.74
        & 95.31 & 28.09 & 36.65
        & \bf 87.28 & 0.69 & 7.63\\
        TAware & 94.38 & 68.33 & 71.86
        & \bf 78.88 & 1.95 & 9.26
        & 92.93 & 26.52 & 30.96
        & \bf 88.07 & 0.35 & 10.63\\
        ~~+ multi & 97.74 & 76.68 & 81.00
        & \bf 95.30 & 4.03 & 10.75
        & 98.31 & 68.87 & 68.22
        & \bf 92.29 & 0.87 & 10.09\\
        KAware & 96.91 & 90.89 & 96.91
        & 64.80 & 13.06 & 7.22
        & 90.93 & 50.92 & 61.08
        & 75.57 & 2.77 & 10.00\\
        Qadpt & 95.65 & 77.33 & 78.68
        & 59.01 & \bf 66.67 & \bf 16.82
        & 92.34 & 38.62 & 36.96
        & 81.24 & \bf 30.85 & \bf 16.87\\
        ~~+ multi & 99.60 & 83.17 & 87.27
        & 56.11 & \bf 61.92 & \bf 18.54
        & 98.47 & 48.78 & 63.54
        & 86.97 & \bf 26.17 & \bf 17.31\\
        ~~+ TAware & 99.02 & 83.14 & 85.59
        & 58.82 & \bf 64.12 & \bf 14.90
        & 98.45 & 56.77 & 65.25
        & 82.52 & \bf 28.34 & \bf 17.68\\\hline
    \end{tabular}
    \caption{The results of change rate and accurate change rate.}
    \label{tab:all-changed}
\end{table*}

\begin{table*}[t!]\small
    \centering
    \begin{tabular}{l|c|c|cZ|c|cZ|c|c|cZ|c|cZ}\hline
        \multirow{3}{*}{\bf Model} 
        & \multicolumn{7}{c|}{\bf HGZHZ}
        & \multicolumn{7}{c}{\bf Friends}\\\cline{2-15}
        & \bf KW & \multicolumn{3}{c|}{\bf KW/Generic} & \multicolumn{3}{c|}{\bf Generated-KW}
        & \bf KW & \multicolumn{3}{c|}{\bf KW/Generic} & \multicolumn{3}{c}{\bf Generated-KW}\\\cline{3-8}\cline{10-15}
         & \bf Acc & \bf Recall & \bf Precision & \bf F1
         & \bf Recall & \bf Precision & \bf F1
         & \bf Acc & \bf Recall & \bf Precision & \bf F1
         & \bf Recall & \bf Precision & \bf F1\\\hline
        Seq2Seq & 12.10 & 29.08 & 27.44 & 28.24
         & 13.30 & 24.28 & 17.18
         & 3.81 & 23.22 & 5.57 & 8.98
         & 6.88 & 2.02 & 3.13\\
        MemNet & 22.58 & 39.09 & 33.41 & 56.21
         & 39.52 & 67.10 & 49.75
         & 22.79 & 37.18 & 24.58 & 54.20
         & 46.02 & 53.98 & 49.69\\
        ~~+ multi & 35.20 & 54.49 & 30.15 & 70.54
         & 60.63 & 83.43 & 70.23
         & 34.92 & 47.31 & 22.78 & 64.24
         & 60.54 & 69.46 & 64.69\\
        TAware & 50.21 & 44.40 & 35.50 & 39.45
         & 49.18 & 76.72 & 59.94
         & 62.74 & \bf 50.78 & 22.50 & 31.19
         & 57.84 & 62.83 & 60.23\\
        ~~+ multi & \bf 59.71 & \bf 68.61 & 28.70 & 40.47
         & \bf 70.18 & \bf 85.54 & 77.10
         & \bf 72.96 & 42.98 & 25.74 & 32.20
         & \bf 71.11 & \bf 77.35 & 74.10\\
        KAware & 20.53 & 40.63 & \bf 36.64 & 38.53
         & 24.61 & 43.13 & 31.34
         & 13.52 & 30.76 & \bf 28.42 & 29.55
         & 15.14 & 18.74 & 16.75\\
        Qadpt & \bf 57.61 & 38.24 & 28.31 & 32.53
         & 44.50 & \bf 90.70 & 59.70
         & \bf 74.00 & 41.33 & 25.31 & 31.39
         & 69.30 & \bf 77.30 & 73.08\\
        ~~+ multi & \bf 57.40 & 51.97 & 28.43 & 36.76
         & \bf 64.55 & \bf 91.22 & 75.60
         & \bf 74.44 & 42.81 & 25.01 & 31.58
         & \bf 74.63 & \bf 77.09 & 75.84\\
        ~~+ TAware & \bf 56.24 & 53.68 & 31.03 & 39.33
         & \bf 63.66 & \bf 88.99 & 74.22
         & \bf 73.57 & 47.05 & 25.91 & 33.42
         & \bf 74.52 & \bf 78.56 & 76.49\\\hline
    \end{tabular}
    \caption{The results of knowledge graph entities prediction. }
    \label{tab:all-kb-f1}
    \vspace{-2mm}
\end{table*}

\subsection{Knowledge-Grounded Conversations}
The recent knowledge-grounded conversation models~\cite{sordoni2015neural, ghazvininejad2018knowledge,zhu2017flexible,zhou2018commonsense} generated responses conditioned on conversation history and external knowledge.
\citet{ghazvininejad2018knowledge} used memory networks~\cite{weston2014memory, weston2015towards, sukhbaatar2015end} to attend on external facts, and added the encoded information to the decoding process.
\citet{zhu2017flexible} added a copy mechanism~\cite{gu2016incorporating, merity2016pointer, he2017generating} for improving its performance.
\citet{zhou2018commonsense} presented two-level graph attention mechanisms~\cite{velivckovic2017graph} to produce more informative responses.

For knowledge from unstructured texts, \citet{ghazvininejad2018knowledge} used bag-of-word representations and \citet{long2017knowledge} applied a convolutional neural network to encode the whole texts.
With structured knowledge graphs, \citet{zhu2017flexible} and \citet{zhou2018commonsense} utilized graph embedding methods (e.g., TransE~\cite{bordes2013translating}) to encode each triplet.

The above methods generated responses without explicit relationship to each external knowledge triplet.
Therefore, when a triplet is added or deleted, it is unknown whether their generated responses can change accordingly.
\citet{moon2019opendialkg} recently presented a similar concept, walking on the knowledge graph, for response generation. 
Nonetheless, their purpose is to find explainable path on a large-scaled knowledge graph instead of adaptation with the changed knowledge graphs. Hence, the proposed attention-based graph walker may suffer from the same issue as previous embedding-based methods.

\subsection{Multi-Hop Reasoning}
We leverage multi-hop reasoning~\cite{lao2011random} to allow our model to quickly adapt to dynamic knowledge graphs.
Recently, prior work used convolutional neural network~\cite{toutanova2015representing}, recurrent neural network~\cite{neelakantan2015compositional, das2017chains}, and reinforcement learning~\cite{xiong2017deeppath, das2017go, chen2018variational, shen2018reinforcewalk} to model multi-hop reasoning on knowledge graphs, and has proved this concept useful in link prediction.
These reasoning models, however, have not yet explored on dialogue generation.
The proposed model is the first attempt at adapting conversations via a reasoning procedure.

\section{Experiments}
\label{sec:exp}

For all models, we use gated recurrent unit (GRU) based Seq2Seq models~\cite{cho2014properties, chung2014empirical, vinyals2015neural}.
Both encoder and decoder for HGZHZ are 256 dimension with 1 layer; ones for Friends are 128 dimension with 1 layer.

We benchmark the task, \emph{dynamic knowledge-grounded dialogue generation}, and corpus \emph{DyKgChat} by providing a detailed comparison between the prior conversational models and our proposed model as the preliminary experiments.
We evaluate their capability of quick adaptation by randomized whole, last 1, last 2 reasoning paths as described in Section~\ref{sec:eval-2}.
We evaluate the produced responses by sentence-level BLEU-2~\cite{papineni2002bleu, liu2016not}, perplexity, distinct-n~\cite{li2016diversity}, and our proposed metrics for predicting knowledge entities descrin section~\ref{sec:eval-1}.

Because of the significant data imbalance of \emph{Friends}, we first train on whole training data, and then fine-tune the models using the subset containing knowledge entities.
Early stopping is adopted in all experiments.

\begin{figure*}[t!]
    \centering
    \subfigure[retrieved relation paths distribution of HGZHZ]{
        \includegraphics[width=.48\linewidth, trim={0.5cm 1.0cm 0.5cm 0.5cm},clip]{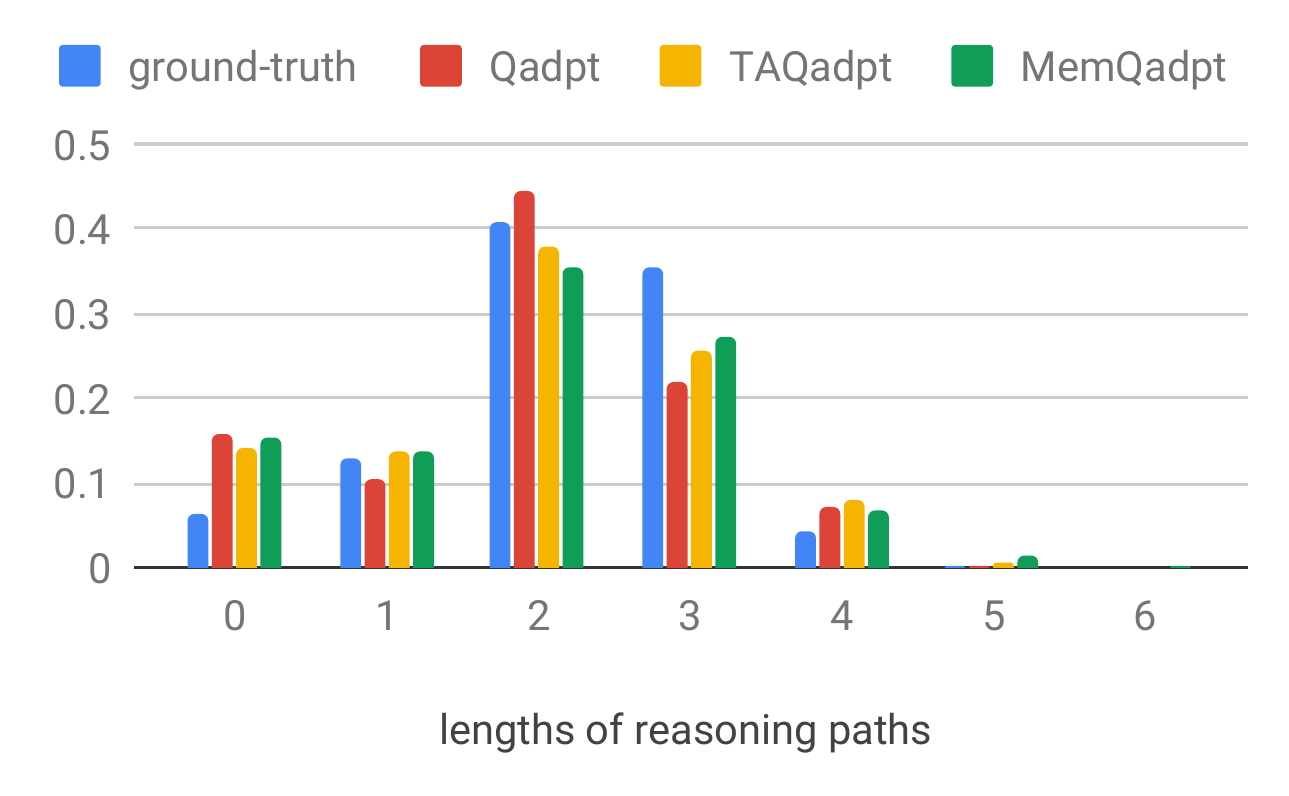}\vspace{-5mm}
    }
    \subfigure[retrieved relation paths distribution of Friends]{
        \includegraphics[width=.48\linewidth, trim={0.5cm 1.0cm 0.5cm 0.5cm},clip]{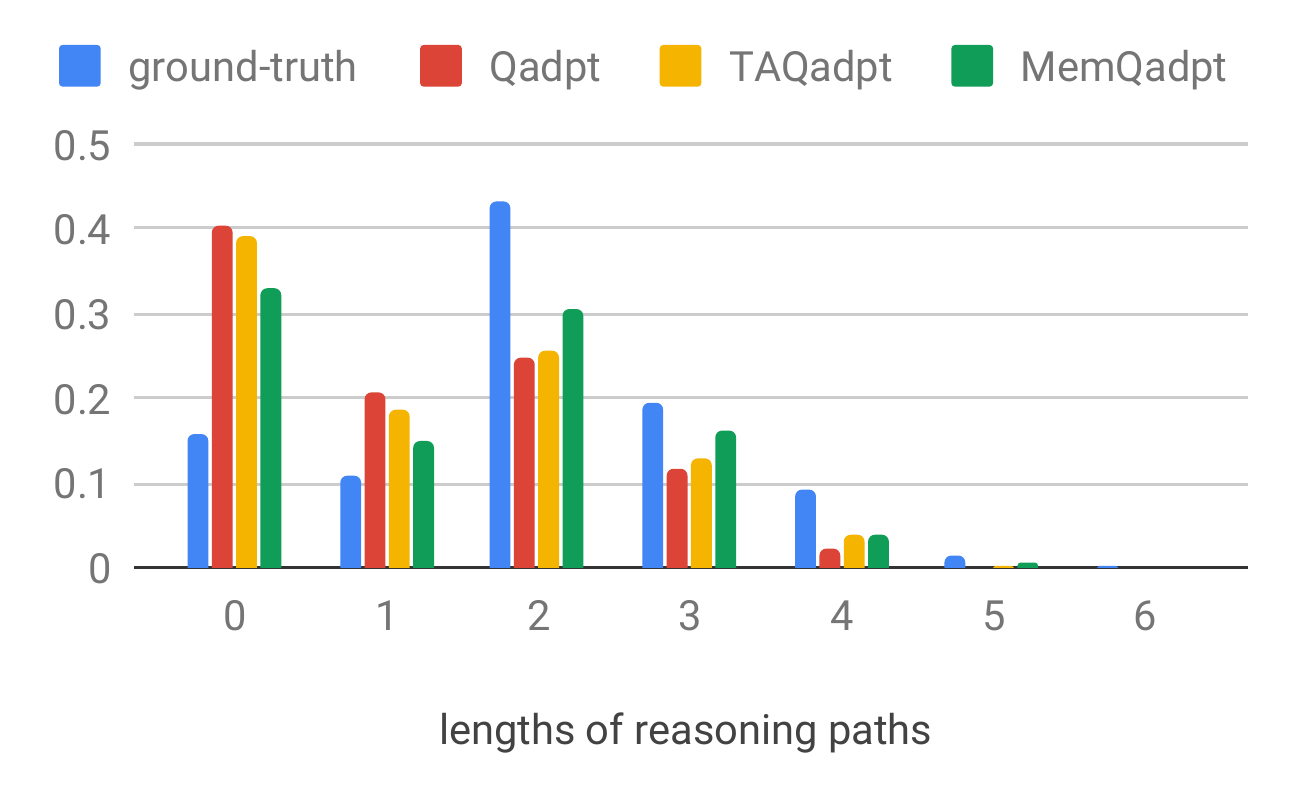}\vspace{-5mm}
    }
    \vspace{-2mm}
    \caption{The distribution of the lengths of Qadpt inferred relation paths.}
    \label{fig:retrieved-paths}
\end{figure*}

\begin{table*}[t!]\small
    \centering
    \begin{tabular}{l|cZ|c|c|c|c|c|cZ|c|c|c|c|c}\hline
        \multirow{2}{*}{\bf Model}
        & \multicolumn{7}{c|}{\bf HGZHZ}
        & \multicolumn{7}{c}{\bf Friends}\\\cline{2-15}
        & \bf BLEU & & \bf PPL & \bf dist-1 & \bf dist-2 & \bf dist-3 & \bf dist-4
        & \bf BLEU & & \bf PPL & \bf dist-1 & \bf dist-2 & \bf dist-3 & \bf dist-4\\\hline
        Seq2Seq & 14.20 & 6.37 & 94.48 & 0.008 & 0.039 & 0.092 & 0.150
        & 15.46 & 4.51 & 73.23 & 0.004 & 0.016 & 0.026 & 0.032\\
        MemNet & \bf 15.73 & 3.96 & 88.29 & 0.012 & 0.062 & 0.150 & 0.240
        & 14.61 & 8.23 & 67.58 & 0.005 & \bf 0.023 & 0.040 & 0.049\\
        + multi & 15.88 & 6.38 & 86.76 & 0.010 & 0.058 & 0.138 & 0.224
        & 12.97 & 3.69 & \bf 54.67 & \bf 0.006 & 0.022 & 0.032 & 0.036\\
        TAware & 15.07 & 1.97 & 81.54 & 0.013 & 0.068 & 0.153 & 0.223
        & 14.78 & 22.25 & 60.61 & 0.002 & 0.007 & 0.013 & 0.016\\
        + multi & 13.34 & 1.00 & \bf 80.48 & \bf 0.022 & 0.122 & 0.239 & 0.304
        & 15.74 & 21.68 & 56.67 & 0.003 & 0.011 & 0.019 & 0.023\\
        KAware & 14.14 & 3.16 & 90.11 & 0.011 & 0.061 & 0.135 & 0.198
        & 15.70 & 20.24 & 64.70 & 0.002 & 0.009 & 0.017 & 0.021\\
        Qadpt & 14.52 & 5.39 & 88.24 & 0.013 & 0.081 & 0.169 & 0.242
        & \bf 17.01 & 28.13 & 68.27 & 0.002 & 0.008 & 0.013 & 0.016\\
        + multi & 15.47 & 1.81 & 86.65 & 0.021 & \bf 0.129 & \bf 0.259 & \bf 0.342
        & 14.79 & 10.85 & 66.70 & 0.005 & \bf 0.023 & \bf 0.041 & \bf 0.051\\
        + TAware & 15.05 & 2.01 & 81.75 & 0.022 & 0.123 & 0.246 & 0.332
        & 16.85 & 22.67 & 55.46 & 0.003 & 0.012 & 0.020 & 0.024\\\hline
    \end{tabular}
    \caption{The results of responses generation with BLEU, perplexity (PPL), distinct scores (1-gram to 4-gram).}
    \label{tab:all-bleu-ppx}
    \vspace{-3mm}
\end{table*}

\subsection{Baselines}
We compare our model with prior knowledge-grounded conversation models: the memory network~\cite{ghazvininejad2018knowledge} and knowledge-aware model (KAware)~\cite{zhu2017flexible, zhou2018commonsense}.
We also leverage the topic-aware model (TAware)~\cite{xing2017topic, wu2018neural, zhou2018emotional} by attending on knowledge graphs and using two separate output projection layers (generic words and all knowledge graph entities).
In our experiments, MemNet is modified for fair comparison, where the memory pool of MemNet stores TransE embeddings of knowledge triples~\cite{zhou2018commonsense}.
The maximum number of the stored triplets are set to the maximum size of all knowledge graphs for each dataset ($176$ for hgzhz and $98$ for friends).
The multi-hop version of MemNet~\cite{weston2014memory} is also implemented (MemNet+multi)\footnote{Note that \emph{multi-hop} here indicates re-attention on the triplet embeddings of a knowledge graph.}.
To empirically achieve better performance, we also utilize the attention of MemNet for TAware and KAware.
Moreover, we empirically find that multi-hop MemNet deteriorate the performance of KAware (compared to one-hop), while it could enhance the performance of TAware.
A standard Seq2Seq model~\cite{vinyals2015neural} is also shown as a baseline without using external knowledge.
We also leverage multi-hop MemNet and the attention of TAware to strength Qadpt (+multi and +TAware).

\begin{table*}[t!]\small
    \centering
    \begin{tabular}{l|ccc|ccc|ccc|ccc}\hline
        \multirow{3}{*}{\bf Model} 
         & \multicolumn{6}{c|}{\bf HGZHZ} & \multicolumn{6}{c}{\bf Friends}\\\cline{2-13}
         & \multicolumn{3}{c|}{\bf Fluency} & \multicolumn{3}{c|}{\bf Information} & \multicolumn{3}{c|}{\bf Fluency} & \multicolumn{3}{c}{\bf Information}\\\cline{2-13}
         & Win & Lose & Kappa & Win & Lose & Kappa & Win & Lose & Kappa &  Win & Lose & Kappa\\\hline
        Qadpt vs Seq2Seq & \bf 60.0 & 21.3 & .69 & \bf 46.0 & 11.3 & .64 & \bf 55.3 & 23.3 & .55 & \bf 60.0 & 13.3 & .61\\
        Qadpt vs MemNet & \bf 49.3 & 19.3 & .78 & \bf 34.0 & 12.7 & .73 & 35.3 & \bf 38.7 & .62 & 17.3 & \bf 22.7 & .59\\
        Qadpt vs TAware & \bf 48.7 & 18.0 & .72 & \bf 30.0 & 13.3 & .70 & \bf 42.7 & 32.7 & .58 & \bf 20.7 & 18.0 & .62\\
        Qadpt vs KAware & \bf 59.3 & 14.0 & .71 & \bf 58.7 & 8.0 & .78 & \bf 44.7 & 28.7 & .62 & \bf 44.7 & 13.3 & .68\\\hline
    \end{tabular}
    \caption{The results of human evaluation.}
    \label{tab:human-eval}
    \vspace{-3mm}
\end{table*}

\subsection{Results}

As shown in Table~\ref{tab:all-changed}, MemNet, TAware and KAware significantly change when the knowledge graphs are largely updated (\emph{All}) and can also achieve good accurate change rate.
For them, the more parts updated (\emph{All} $>>$ \emph{Last2} $>$ \emph{Last1}), the more changes and accurate changes.
However, when the knowledge graphs are slightly updated (\emph{Last1} and \emph{Last2}), the portion of accurate changes over total changes (e.g., the \emph{Last1} score $1.17/31.78$ for HGZHZ with MemNet model) is significantly low.
Among the baselines, KAware has better performance on \emph{Last1}.
On the other hand,
Qadpt outperforms all baselines when the knowledge graphs slightly change  (\emph{Last1} and \emph{Last2}) in terms of accurate change rate.
The proportion of accurate changes over total changes also show significantly better performance than the prior models.
Figure~\ref{fig:retrieved-paths} shows the distribution of lengths of the inferred relation paths for Qadpt models.
After combining TAware or MemNet, the distribution becomes more similar to the test data.

Table~\ref{tab:all-kb-f1} shows the results of the proposed metrics for correctly predicting knowledge graph entities.
On both HGZHZ and Friends, knowledge graph integrated methods demonstrate their power when comparing with Seq2Seq.
We can also observe that although TAware+multi is overall better than the other models for KW/Generic, Qadpt significantly outperforms other baselines on KW-Acc and Generated-KW and performs comparably to TAware+multi.

To evaluate the sentence quality, Table~\ref{tab:all-bleu-ppx} presents the BLEU-2 scores (as recommended in the prior work~\cite{liu2016not}), perplexity (PPL), and distinct-n scores.
The results show that all models have similar levels of BLEU-2 and PPL, while Qadpt+multi has better distinct-n scores.
Even though BLEU-2 is not perfect for measuring the quality of dialogue generation~\cite{liu2016not}, we report it for reference.
Overall, the composition of BLEU-2, PPL and distinct-n shows that the quality of the generated responses of these models are not extremely distinguishable.

\subsection{Human Evaluation}
\label{sec:human-eval}

When performing human evaluation, we randomly select examples from the knowledge entities included outputs of all models (therefore this subsection is not used to evaluate generic responses), because it is difficult for human to distinguish which generic response is better (where in most cases, every one is not proper).
We recruit fifteen annotators to judge the results.
Each annotator was randomly assigned with 20 examples, and was guided to rank the generated responses of the five models: Seq2Seq, MemNet, TAware, KAware, and Qadpt.
They were asked to rank the responses according to the following two criteria: (1) fluency and (2) information.
\emph{Fluency} measures which output is more proper as a response to a given input message.
\emph{Information} measures which output contains more correct information (in terms of knowledge words here) according to a given input message and a referred response.
The evaluation results are classified into ``win'', ``tie'', and ``lose'' for comparison.

The human evaluation results and the annotator agreement in the form of Cohen's kappa~\cite{cohen1960coefficient} are reported in Table~\ref{tab:human-eval}.
According to a magnitude guideline~\cite{landis1977measurement}, most agreements are substantial (0.6-0.8), while some agreements of Friends are moderate (0.4-0.6).
In most cases of Table~\ref{tab:human-eval}, Qadpt outperforms other four models.
However, in Friends, Qadpt, MemNet, and TAware tie closely.
The reason might be the lower annotators agreements of Friends, or only the similar trend with automatic evaluation metrics.
There are two extra spots.
First, Qadpt wins MemNet and TAware less times than winning Seq2Seq and KAware, which aligns with Table~\ref{tab:all-kb-f1} and Table~\ref{tab:all-bleu-ppx}, where MemNet and TAware show better performance than other baselines.
Second, Qadpt wins the baselines more often by fluency than by information, and much more ties happen in the information fields than the fluency fields. This is probably due to the selection of knowledge-contained examples. Hence there is not much difference when seeing the information amount of models.
Overall, the human evaluation results can be considered as reference because of the substantial agreement among annotators and the similar trend with automatic evaluation.

\section{Discussion}

The results demonstrate that MemNet, TAware and Qadpt generally perform better than the other two baselines, and they excel at different aspects.
While Qadpt generally performs the best in most metrics, TAware shows its power in knowledge graph entitiies prediction, and MemNet demonstrates its potential in human evaluation. 

Specifically for the scenario of zero-shot adaptation,
MemNet and TAware show their ability to update responses when the knowledge graphs are largely changed.
On the other hand, Qadpt is better to capture minor dynamic changes (\emph{Last1} and \emph{Last2}) and updates the responses according to the new knowledge graphs\footnote{examples are given in Appendix.}.
This demonstrates that Qadpt has the potential to be explore more, while MemNet and TAware attend on the whole graph instead of focusing on the most influential part.

\section{Conclusion}
This paper presents a new task, \emph{dynamic knowledge-grounded conversation generation}, and a new dataset \emph{DyKgChat} for evaluation.
The dataset is currently provided with a Chinese and an English TV series as well as their correspondent knowledge graphs.
This paper also benchmarks the task and dataset by proposing automatic evaluation metrics and baseline models, which can motivate the future research directions.

\section*{Acknowledgements}
We would like to thank reviewers for their insightful comments. This work was financially supported from the Young Scholar Fellowship Program by Ministry of Science and Technology (MOST) in Taiwan, under Grant 108-2636-E002-003.

\bibliography{emnlp-ijcnlp-2019}
\bibliographystyle{acl_natbib}

\clearpage
\appendix

\begin{CJK*}{UTF8}{bkai}
\begin{table*}[t!]\small
    \centering
    \begin{tabular}{lp{110mm}}
    \hline
        Message Input & 甄嬛（甘露寺，夜。正睡覺，有野貓闖入）: 怎 麼 了 ？ — — 有 貓 ！ \\
        & (Zhen Huan: What's wrong? -- A cat!)\\
        Reference Output & 浣碧（一陣慌亂）: {\bf 槿汐}，快點去打貓，小 姐被貓傷過，見不得貓的！\\
        & Wan-Bi: {\bf Jin-Xi}, go and catch the cat. Mistress was hurt by a cat, and became afraid of it.\\\hline
        Original KG Path & 甄嬛 (Zhen Huan) $\rightarrow$ 主僕 (IsServantOrMasterOf) $\rightarrow$ 槿汐 (Jin-Xi)\\
        Original Output & {\bf 槿汐} ， 是 什 麼 事 啊 ？ ({\bf Jin-Xi}, what happend?)\\
        Changed KG Path & 甄嬛 (Zhen Huan) $\rightarrow$ 主僕 (IsServantOrMasterOf) $\rightarrow$ \textcolor{red}{小允子 (Yun)}\\
        Changed Output & {\bf 小允子} ， {\bf 小允子} ！(Yun, Yun!)\\\hline\hline
        Message Input & [Scene : The restaurant . {\bf Chandler} and {\bf Monica} are sitting at a table]\\ & Monica: Oh my God, he's gonna do it now. Please, I cannot watch this, let's go. \\
        Reference Output & Chandler: I think we have some time. Have you ever heard him talk? \\ & Uh, {\bf Phoebe}, uh, I would be honoured, uh... Spit it out, {\bf David}!\\\hline
        Original KG Path & Phoebe $\rightarrow$ lover $\rightarrow$ David\\
        Original Output & {\bf David} is , {\bf David} , is it ?\\
        Changed KG Path & Phoebe $\rightarrow$ lover $\rightarrow$ \textcolor{red}{Mike}\\
        Changed Output & {\bf Mike}?\\
        \hline
    \end{tabular}
    \caption{Some examples of the response generation with a changed knowledge graph. }
    \label{tab:change-examples}
\end{table*}
\end{CJK*}

\section{Qualitative Analysis}
\label{sec:analysis}
Because the models generate different responses with different knowledge given the same input message, it is difficult to design a universal changed knowledge graph for fair comparison of all models.
We provide the qualitative analysis for further investigation.
Some examples of our proposed Qadpt model with changed knowledge graphs are shown in Table~\ref{tab:change-examples}.
This demonstrates that Qadpt can successfully modify the response with a minor-changed knowledge graph.
More examples of response generation are given in Table~\ref{tab:examples}.

\begin{CJK*}{UTF8}{bkai}
\begin{table*}[t!]\small
    \centering
    \begin{tabular}{lp{130mm}}
    \hline
        Input & 皇帝（眉莊氣急站立不穩，甄嬛忙扶住）：無 妨 ！ 都 一 樣 是 {\bf 太醫} 。 \\
        & (Emperor: It's okay! They are all {\bf imperial doctors}.)\\
        Reference & 皇后（章彌把脈良久）：章 {\bf 太醫} ， 究 竟 什 麼 個 情 形 ？ 莫 非 驚 動 了 胎 氣 ？\\
        & (Empress: {\bf Imperial doctor} Zhang, what's the situation? is it a miscarriage?)\\\hline
        Seq2Seq & {\bf 皇上} ？ ({\bf Emperor}?)\\
        MemNet & {\bf 浣碧} ， 你 看 看 ， 你 看 看 ， 你 是 個 {\bf 太醫} ， 你 也 不 知 道 了 。\\ & (Wan-Bi, you see, you see, you are an {\bf imperial doctor}, and you don't know either.\\
        TAware & 謝 {\bf 皇后} {\bf 娘娘} 。 (Thanks to {\bf Empress} {\bf Madame}.)\\
        KAware & {\bf 妃} {\bf 浣碧} ， 我 都 是 了 ， 我 是 我 的 ， 我 是 我 的 身 子 ， 我 也 是 不 是 。 \\
        & ({\bf Consort} {\bf Wan-Bi}, I am, I'm mine, I'm my body, I am, right?)\\
        Qadpt & {\bf 太醫} ， 這 是 什 麼 ？ ({\bf Imperial Doctor}, what is that?)\\\hline\hline
        Input & 年妃（江福海乘船到湖邊）：你 說 口 諭 便 是 口 諭 了 ？ 等 本 宮 見 了 {\bf 皇上} 再 說 。 \\
        & Consort Nian: Whatever you said. Let me see the {\bf emperor}.\\
        Reference & 江福海：{\bf 娘娘} ， 奴 才 怎 敢 假 傳 {\bf 皇上} 口 諭 呢 ？ {\bf 皇上} 不 想 見 您 也 不 是 一 日 兩 日 了 ，\\
        & (Jiang, Fuhai: {\bf Madame}, I do not full you. Dare I say, the {\bf emperor} do not want to see you.)\\\hline
        Seq2Seq & {\bf 娘娘} ， 您 是 不 是 ？ ({\bf Madame}, you are?)\\
        MemNet & 是 ， {\bf 娘娘} 說 來 了 。 (Yes, {\bf madame} said she is coming.)\\
        TAware & 謝 {\bf 皇上} 。(Thanks to the {\bf emperor}.)\\
        KAware & 奴 婢 給 {\bf 娘娘} 請 安 。 (I pay respects to {\bf madame}.)\\
        Qadpt & 是 ， {\bf 娘娘} ， 您 的 心 思 是 — — (Sure, {\bf madame}, you are thinking about --)\\\hline\hline
        Input & 華妃（園中）：免 了 ， 本 宮 可 不 敢 當 {\bf 莞嬪} 此 禮 。 你 越 是 恭 順 ， 本 宮 就 越 是 覺 得 可 憎 。\\
        & (Consort Hua: It's okay, I don't want you pay respects to me. The more respectful\\ & you are, the more horror I feel.)\\
        Reference & 甄嬛：{\bf 娘娘} 真 是 說 笑 了 ， 難 道 {\bf 娘娘} 是 喜 歡 妹 妹 對 {\bf 娘娘} 不 恭 不 順 直 言 犯 上 嗎 ？ 妹 妹 並 不 敢 肆 意 冒 犯 {\bf 娘娘} 。\\
        & (Zhen Huan: {\bf Madame} is kidding. Does {\bf madame} want me to say the truth? I'm not dare to offend you.)\\\hline
        Seq2Seq & {\bf 臣妾} 不 敢 。(I dare not do it.)\\
        MemNet & {\bf 娘娘} ， {\bf 臣妾} 不 敢 。({\bf Madame}, I dare not do it.)\\
        TAware & {\bf 娘娘} 還 是 這 樣 ， {\bf 娘娘} 也 不 是 不 是 {\bf 娘娘} 的 。 ({\bf Madame} is still the same. {\bf Madame} is not {\bf madame}'s)\\
        KAware & 謝 {\bf 妃} {\bf 朕} 。 (Thanks to {\bf Consort} I.)\\
        Qadpt & {\bf 娘娘} ， {\bf 娘娘} 別 多 心 了 。 ({\bf Madame}, {\bf Madame}, don't think too much.)\\\hline\hline
        Input & [Scene : The Rehearsal Dinner, {\bf Chandler} and {\bf Monica} are greeting guests as they arrive. A woman enters.] Mr. geller: So are you his mother or his father?\\
        Reference & Mrs. geller: {\bf Jack}!\\\hline
        Seq2Seq & {\bf Ross} , I'm so sorry.\\
        MemNet & Are you guys doing {\bf Chandler}?\\
        TAware & Well, I'm sorry, I'm sorry, I'm sorry, I'm sorry, {\bf Chandler}.\\
        KAware & {\bf Ross} .\\
        Qadpt & {\bf Jack}'s gonna be here!\\\hline\hline
        Input & [Scene: The Rehearsal Dinner, {\bf Chandler} and {\bf Monica} are greeting guests as they arrive. A woman enters.] Ross: Can I have everyone's attention please? I'm uh; I'm {\bf Ross\_Geller}.\\
        Reference & Mr. geller: {\bf Doctor} {\bf Ross\_Geller} .\\\hline
        Seq2Seq & Hey, {\bf Joey}, you're gonna have a baby.\\
        MemNet & Well, I'm sorry, I'm not going to be with {\bf Emma}.\\
        TAware & Oh, {\bf Rachel}.\\
        KAware & Hey, {\bf Ross}, you're a great guy.\\
        Qadpt & Hey, {\bf Ross\_Geller}, I'm gonna go.\\\hline\hline
        Input & [Scene: {\bf Monica} and {\bf Rachel}'s, {\bf Chandler} and {\bf Monica} and {\bf Rachel} and {\bf Phoebe} are planning \\ & their respective strategies to break the other pairing. {\bf Joey} is not amused.] Joey: Hey, check it out! Check it out! Ugly Naked Guy has a naked friend!\\
        Reference & Rachel: Oh yeah! Oh my God! That is our friend! It's Naked {\bf Ross}!\\\hline
        Seq2Seq & Hey, {\bf Monica}, I'm sorry.\\
        MemNet & Hey {\bf Ross}!\\
        TAware & {\bf Ross}!\\
        KAware & Oh, {\bf Ross}, you're so sorry.\\
        Qadpt & {\bf Ross}, you have to go to the bathroom!\\
        \hline
    \end{tabular}
    \caption{Some examples of the response generation. }
    \label{tab:examples}
\end{table*}
\end{CJK*}

\end{document}